\documentclass[10pt,twocolumn,letterpaper]{article}

\usepackage{iccv}
\usepackage{times}
\usepackage{epsfig}
\usepackage{graphicx}
\usepackage{amsmath}
\usepackage{amssymb}
\usepackage{cite}
\usepackage{subfigure} 
\usepackage{enumitem}
\usepackage{colortbl}
\usepackage[table,xcdraw]{xcolor}
\usepackage[breaklinks=true,bookmarks=false]{hyperref}

\iccvfinalcopy % *** Uncomment this line for the final submission

\def\iccvPaperID{****} % *** Enter the ICCV Paper ID here

\ificcvfinal\fi

\begin{document}

%%%%%%%%% TITLE
\title{Predicting the Future: A Jointly Learnt Model for Action Anticipation}

\author{Harshala Gammulle \hspace{5mm} Simon Denman \hspace{5mm} Sridha Sridharan \hspace{5mm} Clinton Fookes \\
Image and Video Research Lab, SAIVT, Queensland University of Technology (QUT), Australia\\
{\tt\small  \{pranali.gammule, s.denman, s.sridharan, c.fookes\}@qut.edu.au}
}

\maketitle
\ificcvfinal\thispagestyle{empty}\fi

%%%%%%%%% ABSTRACT
\begin{abstract}
Inspired by human neurological structures for action anticipation, we present an action anticipation model that enables the prediction of plausible future actions by forecasting both the visual and temporal future. In contrast to current state-of-the-art methods which first learn a model to predict future video features and then perform action anticipation using these features, the proposed framework jointly learns to perform the two tasks, future visual and temporal representation synthesis, and early action anticipation. The joint learning framework ensures that the predicted future embeddings are informative to the action anticipation task. Furthermore, through extensive experimental evaluations we demonstrate the utility of using both visual and temporal semantics of the scene, and illustrate how this representation synthesis could be achieved through a recurrent Generative Adversarial Network (GAN) framework. Our model outperforms the current state-of-the-art methods on multiple datasets: UCF101, UCF101-24, UT-Interaction and TV Human Interaction. \footnote{This research was supported by an Australian Research Council (ARC) Linkage grant LP140100221}
\end{abstract}

%%%%%%%%% BODY TEXT
\section{Introduction}

\begin{figure}[htbp]
        \centering
       % \vskip -5pt
       \subfigure[Action Recognition]{\includegraphics[width=0.95\linewidth]{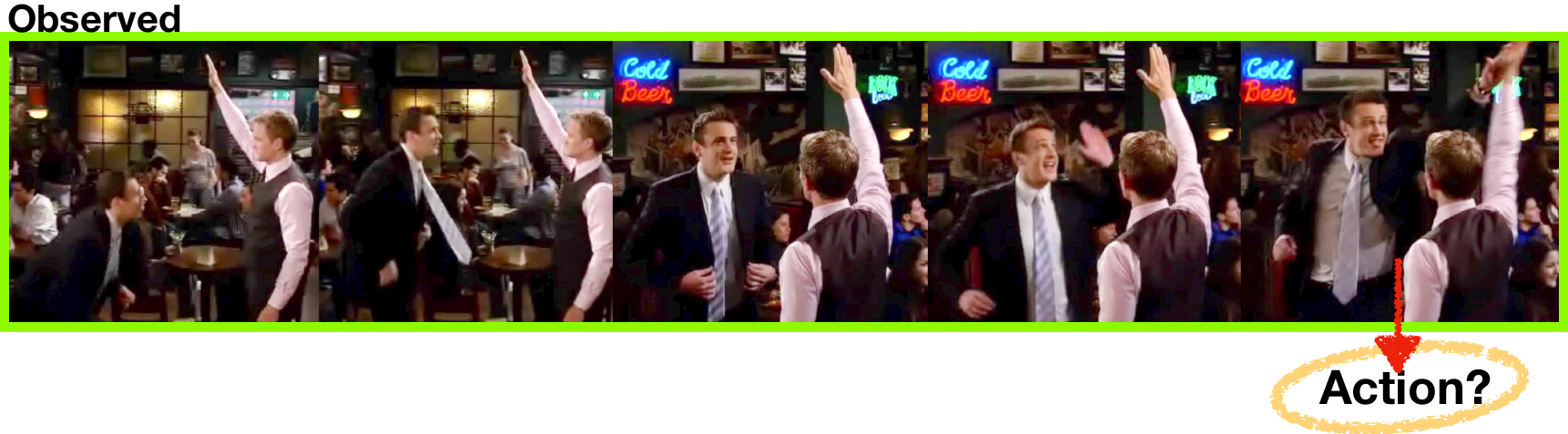}}  \vskip -5pt
	\subfigure[Typical Action Anticipation]{\includegraphics[width=0.95\linewidth]{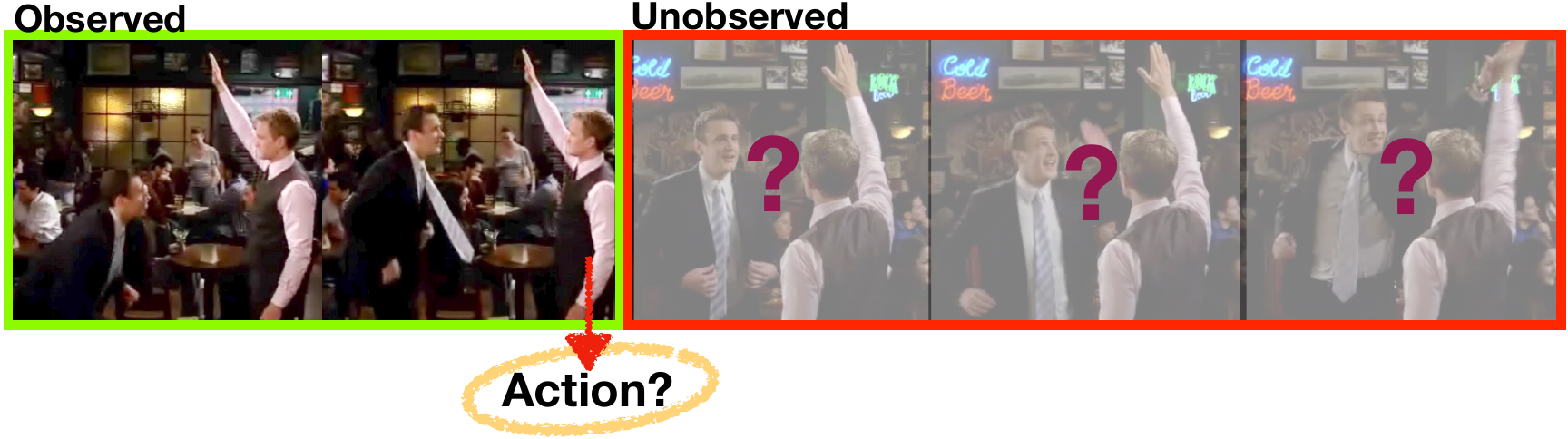}} \vskip -2pt
	\subfigure[Proposed Action Anticipation Method]{\includegraphics[width=0.95\linewidth]{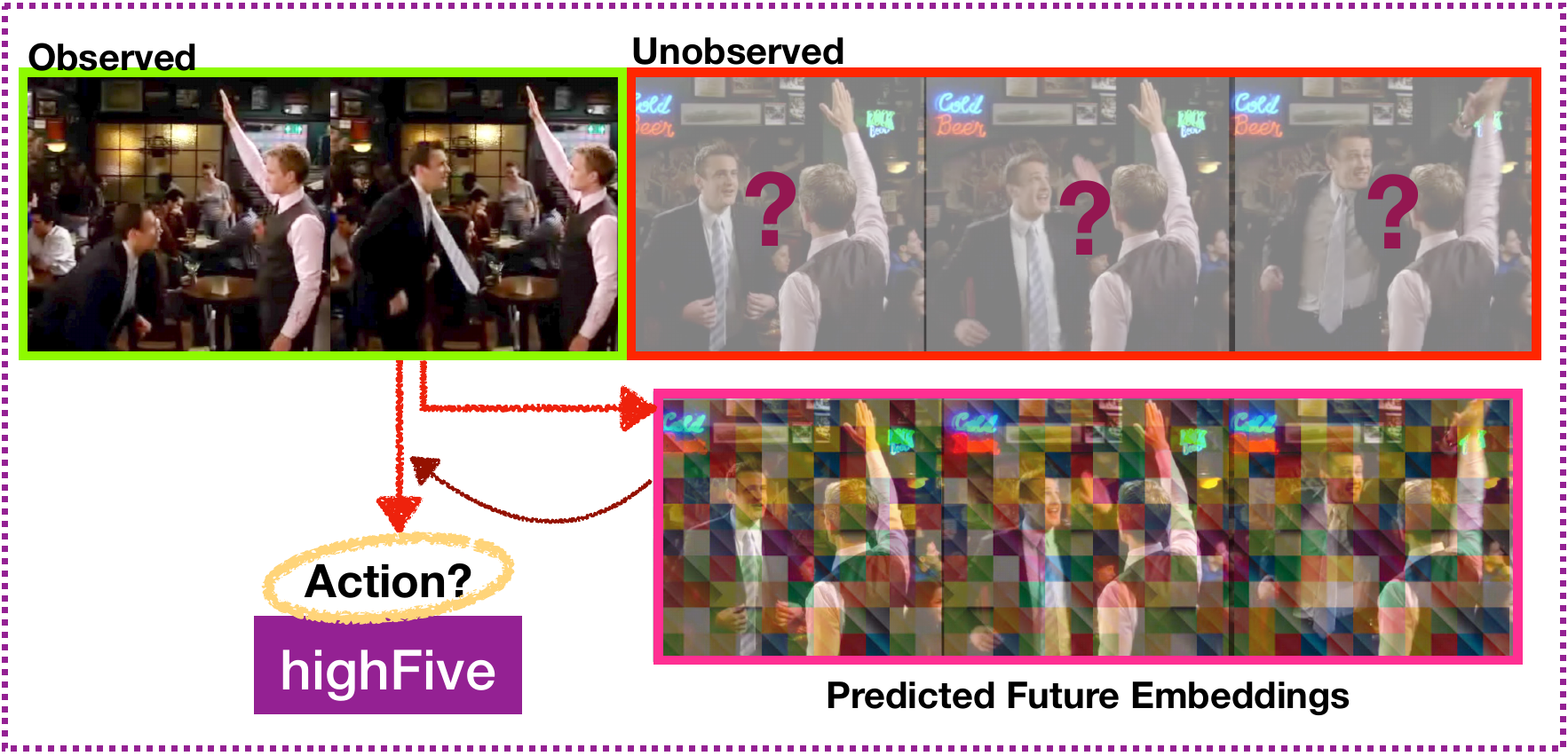}}
	\caption{Action anticipation through future embedding prediction. Action recognition approaches (a) carry out the recognition task via fully observed video sequences while the typical action anticipation methods (b) are based on predicting the action from a small portion of the frames. In our proposed model (c) we jointly learn the future frame embeddings to support the anticipation task.}
	\vspace{-4mm}
	\label{fig:front_fig}
\end{figure}

We propose an action anticipation model that uses visual and temporal data to predict future behaviour, while also predicting a frame-wise future representation to support the learning. Unlike action recognition where the recognition is carried out after the event, by observing the full video sequence (Fig. \ref{fig:front_fig}(a)), the aim of action anticipation (Fig. \ref{fig:front_fig}(b)) is to predict the future action as early as possible by observing only a portion of the action \cite{mohammad_iccv2017}. Therefore, for the prediction we only have partial information in the form of a small number of frames, so the available information is scarce. Fig. \ref{fig:front_fig}(c) shows the intuition behind our proposed model. The action anticipation task is accomplished by jointly learning to predict the future embeddings (both visual and temporal) along with the action anticipation task, where the anticipation task provides cues to help compensate for the missing information from the unobserved frame features. We demonstrate that joint learning of the two tasks complements each other. 

This approach is inspired by recent theories of how humans achieve the action predictive ability. Recent psychology literature has shown that humans build a mental image of the future, including future actions and interactions (such as interactions between objects) before initiating muscle movements or motor controls \cite{medical_1,medical_2,medical_3}. These representations capture both the visual and temporal information of the expected future. Mimicking this biological process, our action anticipation method jointly learns to anticipate future scene representations while predicting the future action, and outperforms current state-of-the-art methods.

\begin{figure*}[htbp]
        \centering
        	\includegraphics[width=0.95\linewidth]{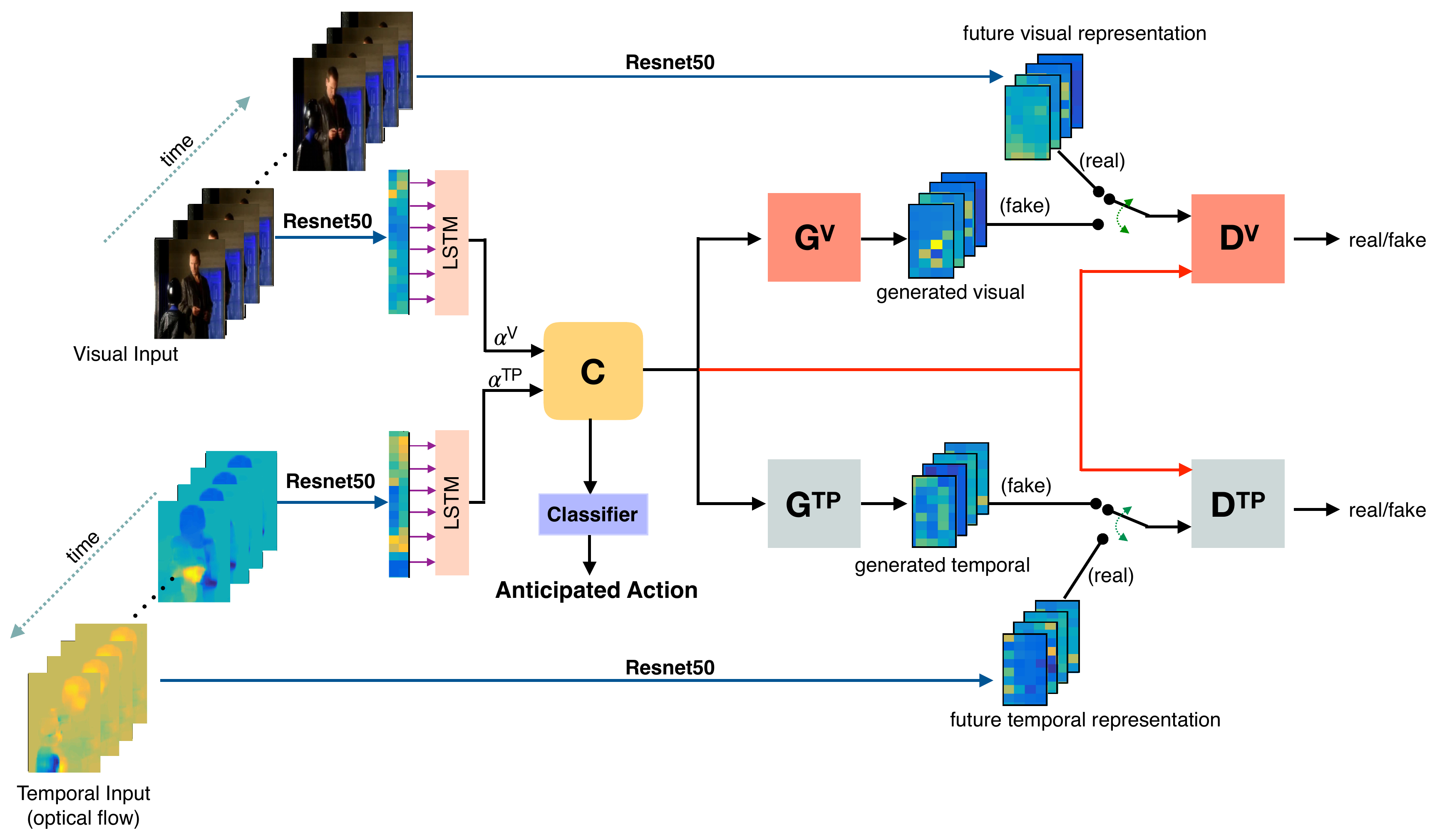}
	\caption{Action Anticipation GAN (AA-GAN): The model receives RGB and optical flow streams as the visual and temporal representations of the given scene. Rather than utilising the raw streams we extract the semantic representation of the individual streams by passing them through a pre-trained feature extractor. These streams are merged via an attention mechanism which embeds these low-level feature representations in a high-level context descriptor. This context representation is utilised by two GANs: one for future visual representation synthesis and one for future temporal representation synthesis; and the anticipated future action is obtained by utilising the context descriptor. Hence context descriptor learning is influenced by both the future representation prediction, and the action anticipation task.}
	\label{fig:multi_gan}
	\vspace{-3mm}
\end{figure*}

In contrast to recent works \cite{vondrickCVPR16,shi2018,mohammad_iccv2017} which rely solely on visual inputs, and inspired by \cite{medical_1,medical_2,medical_3}, we propose a joint learning process which attends to salient components of both visual and temporal streams, and builds a highly informative context descriptor for future action anticipation. In \cite{vondrickCVPR16} the authors demonstrate that the context semantics, which capture high level action related concepts including environmental details, objects, and historical actions and interactions, are more important when anticipating actions than the actual pixel values of future frames. Furthermore, the semantics captured through pre-trained deep learning models show robustness to background and illumination changes as they tend to capture the overall meaning of the input frame rather than simply using pixel values \cite{donahue2013deep,krizhevsky2012imagenet}. Hence in the proposed architecture we extract deep visual and temporal representations from the inputs streams and predict the future representations of those streams.  

Motivated by recent advances in Generative Adversarial Networks (GAN) \cite{GAN2014,GAN_action_1, GAN_action_2} and their ability to automatically learn a task specific loss function, we employ a GAN learning framework in our approach as it provides the capability to predict a plausible future action sequence.

Although there exist individual GAN models for anticipation \cite{zhang2017deep, li2018convolutional}, we take a step further in this work. The main contribution is the joint learning of a context descriptor for two tasks, action anticipation and representation prediction, through the joint training of two GANs. 

Fig. \ref{fig:multi_gan} shows the architecture of our proposed Action Anticipation GAN (AA-GAN) model. The model receives the video frames and optical flow streams as the visual and temporal representations of the scene. We extract a semantic representation of the individual streams by passing them through a pre-trained feature extractor, and fuse them through an attention mechanism. This allows us to provide a varying level of focus to each stream and effectively embed the vital components for different action categories. Through this process low level feature representations are mapped to a high-level context descriptor which is then used by both the future representation synthesis and classification procedures. By coupling the GANs (visual and temporal synthesisers) through a common context descriptor, we optimally utilise all available information and learn a descriptor which better describes the given scene.
       
Our main contributions are as follow: 

\begin{itemize}
  \item We propose a joint learning framework for early action anticipation and synthesis of the future representations.
  \item We demonstrate how attention can efficiently determine the salient components from the multi-modal information, and generate a single context descriptor which is informative for both tasks. 
  \item We introduce a novel regularisation method based on the exponential cosine distance, which effectively guides the generator networks in the prediction task.
    \item We perform evaluations on several challenging datasets, and through a thorough ablation study, demonstrate the relative importance of each component of the proposed model.
\end{itemize}

\section{Previous Work}
\label{sec:related_work}

Human action recognition is an active research area that has great importance in multiple domains \cite{application_1,application_2,application_3}. Since the inception of the field researchers have focused on improving the applicability of methods to tally with real world scenarios. The aim of early works was to develop discrete action recognition methods using image \cite{imageBased1,imageBased2} or video inputs \cite{actiontubes, twoStream,3dcnn}, and these have been extended to detect actions in fine-grained videos \cite{leaCVPR,Ni2016}. Although these methods have shown impressive performance, they are still limited for real-world applications as they rely on fully completed action sequences. This motivates the development of action anticipation methods, which can accurately predict future actions utilising a limited number of early frames, and thereby providing the ability to predict actions that are in progress.

In \cite{vondrickCVPR16}, a deep network is proposed to predict a representation of the future. The predicted representation is used to classify future actions. However \cite{vondrickCVPR16} requires the progress level of the ongoing action to be provided during testing, limiting applicability \cite{huTPAMI2018}. Hu et al. \cite{huTPAMI2018} introduced a soft regression framework to predict ongoing actions. This method \cite{huTPAMI2018} learns soft labels for regression on the subsequences containing partial action executions. Lee et al. \cite{PR2019_Lee} proposed a human activity representation method, termed sub-volume co-occurrence matrix, and developed a method to predict partially observed actions with the aid of a pre-trained CNN. The deep network approach of Aliakbarian et al. \cite{mohammad_iccv2017} used a multi-stage LSTM architecture that incorporates context-aware and action-aware features to predict classes as early as possible. The CNN based action anticipation model of \cite{Rodriguez2018_eccvw} predicts the most plausible future motion, and was improved via an effective loss function based on dynamic and classification losses. The dynamic loss is obtained through a dynamic image generator trained to generate class specific dynamic images. However, performance is limited due to the hand-crafted loss function. A GAN based model can overcome this limitation as it can automatically learn a loss function and has shown promising performance in recent research \cite{mathieu2015, yoo2016, inpainting}. 
          
%The GAN was introduced by Goodfellow et al. \cite{GAN2014} and has became increasingly popular within a short period of time. 
In our work we utilise a conditional GAN \cite{condGAN,gammulle2018multi,gammulle2019coupled} for deep future representation generation. A limited number of GAN approaches can be found for human action recognition \cite{GAN_action_1, GAN_action_2}. In \cite{GAN_action_1}, a GAN is used to generate masks to detect the actors in input frame and action classification is done via a CNN. This method is prone to difficulties with the loss function as noted previously. Considering other GAN methods, \cite{li2018convolutional, zhang2017deep} require human skeletal data which is not readily available; \cite{zhang2017deep} only synthesises the future skeletal representation; and \cite{zeng2017visual} considers the task of synthesising future gaze points using a single generator and discriminator pair and directly extracting spatio-temporal features from a 3D CNN. In contrast to these, we analyse two modes and utilise an attention mechanism to embed the salient components of each mode into a context descriptor which can be used for multiple tasks; and we learn this descriptor through joint training of two GANs and a classifier. 

The authors of \cite{shi2018} have adapted the model of \cite{vondrickCVPR16} to a GAN setting; using GANs to predict the future visual feature representation. Upon training this representation, they train a classifier on the predicted features to anticipate the future action class. We argue that the approach of \cite{shi2018} is suboptimal, as there is no guarantee that the future action representation is well suited to predicting the action due to the two stage learning. Our approach, which learns the tasks jointly, ensures that a rich multi-modal embedding is learnt that captures the salient information needed for both tasks. Furthermore, by extending this to a multi-modal setting, we demonstrate the importance of attending to both visual and temporal features for the action anticipation task.

\section{Action Anticipation Model}
\label{sec:methods}

Our action anticipation model is designed to predict the future while classifying future actions. The model aims to generate embeddings for future frames, to obtain a complete notion of the ongoing action and to understand how best to classify the action. In Sec. \ref{sec:methods_ct}, we discuss how the context descriptor is generated using the visual and temporal input streams while Sec. \ref{sec:methods_gans} describes the use of the GAN in the descriptor generation process. The future action classification procedure is described in Sec. \ref{sec:methods_classif} and we further improve this process with the addition of the cosine distance based regularisation method presented in Sec. \ref{sec:methods_regular}.
  
\subsection{Context Descriptor Formulation}
\label{sec:methods_ct}

Inputs to our model are two fold: visual and temporal. The visual inputs are the RGB frames and the temporal inputs are the corresponding optical flow images (computed using \cite{brox2004}). If the number of input video frames is T, then both the visual input ($I^V$) and the temporal input ($I^{TP}$) can be represented as follows,

\begin{equation}
\begin{split}
I^V = \{I^V_{1}, I^V_{2},\dots, I^V_{T}\} , \\I^{TP} = \{I^{TP}_{1}, I^{TP}_{2},\dots, I^{TP}_{T}\}.
\end{split}
\label{eq:1}
\end{equation} 

%\begin{equation}
%I^{TP} = \{I^{TP}_{1}, I^{TP}_{2},\dots, I^{TP}_{T}\}.
%\label{eq:1}
%\end{equation} 

These inputs are passed through a pre-trained feature extractor which extracts features $\theta^V$ and $\theta^{TP}$ frame wise, %where $\theta^V$ and $\theta^{TP}$ are given by,
 
\begin{equation}
\begin{split}
\theta^V=\{\theta^V_1, \theta^V_2,\dots,\theta^V_T\}, \\\theta^{TP}=\{\theta^{TP}_1, \theta^{TP}_2,\dots,\theta^{TP}_T\}.  
\end{split}
\label{eq:2}
\end{equation} 

Then $\theta^V$ and $\theta^{TP}$ are sent through separate LSTM networks to capture the temporal structure of the input features. The LSTM outputs are defined as,

\begin{equation}
h^V_{t}=LSTM(\theta^V_{t}), h^{TP}_{t}=LSTM(\theta^{TP}_{t}).
\label{eq:3}
\end{equation} 

Attention values are generated for each frame such that,

\begin{equation}
e^V_{t}= \mathrm{tanh}(a^V[h^V_{t}]^\top), e^{TP}_{t}= \mathrm{tanh}(a^{TP}[h^V_{t}]^\top),
\label{eq:4}
\end{equation} 
where $a^{V}$ and $a^{TP}$ are multilayer perceptrons trained together with the rest of the network, and are passed through a sigmoid function to get the score values,

\begin{equation}
%\alpha^V_{t}=\frac{\mathbb{E}(e^V_{t})} {\sum_{l=1}^{T}\mathbb{E}(e^V_{l})}, \alpha^{TP}_{t}=\frac{\mathbb{E}(e^{TP}_{t})} {\sum_{l=1}^{T}\mathbb{E}(e^{TP}_{l})}.
\alpha^V_{t}=\sigma([e^V_{t}, e^{TP}_{t}]), \alpha^{TP}_{t}=1- \alpha^V_{t}.
\label{eq:5}
\end{equation}

Then, an attention weighted output vector is generated,

\begin{equation}
%\tilde{\mu}^V=\sum_{t=1}^{T} \alpha^V_{t} h^V_{t}, \tilde{\mu}^{TP}=\sum_{t=1}^{T} \alpha^{TP}_{t} h^{TP}_{t}.
\tilde{\mu}^V_{t}=\alpha^V_{t}h^V_{t} , \tilde{\mu}^{TP}_{t}=\alpha^{TP}_{t}h^V_{t}
\label{eq:6}
\end{equation} 

Finally these output vectors are concatenated (denoted by $[, ]$) to generate the context descriptor ($C_{t}$), 

\begin{equation}
C_{t}=[\tilde{\mu}^V_{t}, \tilde{\mu}^{TP}_{t}].
\label{eq:7}
\end{equation} 
$C_{t}$ encodes the recent history of both inputs, and thus is used to predict future behaviour.

\subsection{Visual and Temporal GANs}
\label{sec:methods_gans}

GAN based models are capable of learning an output that is difficult to discriminate from real examples. They learn a mapping from the input to this realistic output while learning a loss function to train the mapping. The context descriptor, $C_{t}$, is the input for both GANs (visual and temporal synthesisers, see Fig. \ref{fig:multi_gan}). The ground truth future visual and temporal frames are denoted $F^{V}$ and $F^{TP}$, and are given by,

\begin{equation}
\begin{split}
F^V=\{F^V_{1},F^V_{2},\dots,F^V_{T}\}, \\ F^{TP}=\{F^{TP}_{1},F^{TP}_{2},\dots,F^{TP}_{T}\}.
\end{split}
\label{eq:8}
\end{equation}

We extract features for $F^V$ and $F^{TP}$ similar to Eq. \ref{eq:2},

\begin{equation}
\begin{split}
\beta^V=\{\beta^V_1, \beta^V_2,\dots,\beta^V_T\}, \\\beta^{TP}=\{\beta^{TP}_1, \beta^{TP}_2,\dots,\beta^{TP}_T\}.  
\end{split}
\label{eq:9}
\end{equation}

These features, $\beta^V$ and $\beta^{TP}$, are utilised during GAN training. The aim of the generator ($G^V$ or $G^{TP}$) of each GAN is to synthesise the future deep feature sequence that is sufficiently realistic to fool the discriminator ($D^V$ or $D^{TP}$). It should be noted that the GAN models do not learn to predict the future frames, but the deep features of the frames (visual or temporal). As observed in \cite{vondrickCVPR16} this allows the model to recognise higher-level concepts in the present and anticipate their relationships with future actions. This is learnt through the following loss functions,

\begin{equation}
\begin{split}
\mathcal{L}^V(G^V, D^V)= \sum_{t=1}^{T} \log D^V(C_{t}, \beta^V) +\\ \sum_{t=1}^{T} \log (1-D^V(C_{t}, G^V(C_{t}))),
\end{split}
\label{eq:10}
\end{equation}

\begin{equation}
\begin{split}
\mathcal{L}^{TP}(G^{TP}, D^{TP})= \sum_{t=1}^{T} \log D^{TP}(C_{t}, \beta^{TP})+\\\sum_{t=1}^{T} \log (1-D^{TP}(C_{t}, G^{TP}(C_{t}))).
\end{split}
\label{eq:11}
\end{equation}

\subsection{Classification}
\label{sec:methods_classif}

The deep future sequences are learnt through the two GAN models as described in Sec. \ref{sec:methods_gans}. A naive way to perform the future action classification is using the trained future feature predictor and passing the synthesised future features to the classifier. However, this is sub-optimal as $G^{V}$ and $G^{TP}$ have no knowledge of this task, and thus features are likely sub-optimal for it. As such, in this work we investigate joint learning of the embedding prediction and future action anticipation, allowing the model to learn the salient features that are required for action anticipation. Hence, the GANs are able to support learning salient features for both processes. We perform future action classification for the action anticipation task through a classifier, the input for which is $C_{t}$. Then the classification loss can be defined as,

\begin{equation}
\begin{split}
\mathcal{L}^{C}= - \sum_{t=1}^{T} y_{t}\log f^C(C_{t}).
\end{split}
\label{eq:12}
\end{equation} 
It is important to note that the context descriptor $C_t$ is influenced by both the classification loss, $L^{c}$, and the GAN losses, $L^V$ and $L^{TP}$, as $G^{V}$ and $G^{TP}$ utilise the context descriptor to synthesise the future representations. 

\subsection{Regularisation}
\label{sec:methods_regular}

To stabilise GAN learning a regularisation method such as the $L_2$ loss is often used \cite{Isola_CVPR2017}. However the cosine distance has been shown to be more effective when comparing deep embeddings \cite{wang2017learning,wojke2018deep}. Furthermore when generating future sequence forecasts it is more challenging to forecast representations in the distant future than the near future. However, the semantics from the distant future are more informative for the action class anticipation problem, as they carry more information about what the agents are likely to do. Hence we propose a temporal regularisation mechanism which compares the predicted embeddings with the ground truth future embeddings using the cosine distance, and encourages the model to focus more on generating accurate embeddings for the distant future, 

\begin{equation}
\begin{split}
\mathcal{L}^{R}= \sum_{t=1}^{T} - e^{t} d(G^V(C_{t}), \beta^V_{t}) + \sum_{t=1}^{T} - e^{t} d(G^{TP}(C_{t}), \beta^{TP}_{t}),
\end{split}
\label{eq:13}
\end{equation} 
where $d$ represents the cosine distance function. Motivated by \cite{mohammad_iccv2017} we introduce the exponential term, $ e^{t}$, encouraging more accurate prediction of distant future embeddings.
       
Then, the loss for the final model that learns the context descriptor $C_{t}$ and is reinforced by both deep future sequence synthesisers (GAN models), and the future action classification can be written as,

\begin{equation}
\begin{split}
\mathcal{L} = w^V \mathcal{L}^V+w^{TP} \mathcal{L}^{TP}+w^c \mathcal{L}^{C}+w^R \mathcal{L}^{R},
\end{split}
\label{eq:14}
\end{equation} 

where $w^V, w^{TP}, w^c$ and $w^R$ are hyper-parameters which control the contribution of the respective losses.          

\section{Evaluations}
\label{sec:evaluate}
\subsection{Datasets}

Related works on action anticipation or early action prediction typically use discrete action datasets. The four datasets we use to evaluate our work are outlined below.

\textbf{UCF101 \cite{UCF101}} has been widely used for discrete action recognition and recent works for action anticipation due to its size and variety. The dataset includes 101 action classes from 13,320 videos with an average length of 7.2 seconds. In order to perform comparison to the state-of-the-art methods, we utilise the provided three training/testing splits and report the average accuracy over three splits.  

\textbf{UCF101-24 \cite{UCF101_24}} is a subset of the UCF101 dataset. It is composed of 24 action classes in 3207 videos. In order to compare action anticipation results to the state-of-the-art we utilise only the data provided in set1.

\textbf{UT-Interaction (UTI) \cite{UTI}} is a human interaction dataset, which contains videos of two or more people performing interactions such as handshake, punch etc. in a sequential and/or concurrent manner. The dataset has total of 120 videos. For the state-of-the-art comparison we utilise a 10-fold leave-one-out cross validation on each set and the mean performance over all sets is obtained, as per \cite{mohammad_iccv2017}.

\textbf{TV Human Interaction (TV-HI) \cite{TV_human}} dataset is a collection of 300 video clips collected from 20 different TV shows. It is composed of four action classes of people performing interactions such as handshake, highfive, hug and kiss, and a fifth action class called `none' which does not contain any of the four actions. The provided train/ test splits are utilised with a 25-fold cross validation, as per \cite{vondrickCVPR16}. 

\subsection{Network Architecture and Training}

Considering related literature for different datasets, different numbers of observed frames \cite{mohammad_iccv2017,shi2018} are used. Let $T$ be the number of observed frames, then we extract frames $T+1$ to $T+\acute{T}$ as future frames, where $\acute{T}$ is the number of future frames for embedding prediction. As the temporal input, similar to \cite{twoStream} we use dense optical flow displacements computed using \cite{brox2004}. In addition to horizontal and vertical components we also use the mean displacement of the horizontal and vertical flow. Both visual and temporal inputs are individually passed through a pre-trained ResNet50 \cite{resnet} trained on ImageNet \cite{imageNet}, and activations from the  `activation\_23'  layer are used as the input feature representation. 

%The architectures of the generator, discriminator and the classifier are presented in Fig. \ref{fig:network_arc}.
The network of the generator is composed of two LSTM layers followed by a fully connected layer. The generator is fed only with the context input while the discriminator is fed with both the context and the real/fake feature representations. The two inputs of the discriminator are passed through separate LSTM layers and then the merged output is passed through two fully connected layers. The classifier is composed of a single LSTM layer followed by a single fully connected layer. For clarity, we provide model diagrams in the supplementary materials. For all LSTMs, 300 hidden units are used. For the model training procedure we follow the approach of \cite{Isola_CVPR2017}, alternating between one gradient decent pass for the discriminators, and the generators and the classifier using 32 samples per mini batch. The Adam optimiser \cite{adam2015} is used with a learning rate of 0.0002 and a decay of $ 8\times{10^{-9}}$, and is trained for 40 epochs. Hyper-parameters $w^V, w^{TP}, w^c, w^R$ are evaluated experimentally and set to 25, 20, 43 and 15, respectively. Please refer to supplementary material for these evaluations.  When training the proposed model for the UTI and TV-HI datasets, due to the limited availability of training examples we first train the model on UCF101 training data and fine-tuned it on the training data from the specific datasets.

For the implementation of our proposed method we utilised Keras \cite{keras} with Theano \cite{theano} as the backend. 
%\begin{figure*}[htbp]
%  \centering
%        \subfigure[$G^V/ G^{TP}$]{\includegraphics[width=0.25\textwidth]{Figures/G.pdf}}
%    	\subfigure[$D^V/ D^{TP}$]{\includegraphics[width=0.3\textwidth]{Figures/D.pdf}}
%	\subfigure[Classifier]{\includegraphics[width=0.25\textwidth]{Figures/classif.pdf}}
%   \caption{The architectures of generator ($G^V/ G^{TP}$), discriminator ($D^V/ D^{TP}$) and the classifier. }
%  \label{fig:network_arc}
%\end{figure*}

\subsection{Performance Evaluation}

\subsubsection{Evaluation Protocol}

To evaluate our model on each dataset, where possible we consider two settings for the number of input frames, namely the `Earliest' and `Latest' settings. For UCF101 and UTI, similar to \cite{mohammad_iccv2017} we consider 20\% and 50\% of the frames for the `Earliest' and `Latest' settings, respectively; following \cite{mohammad_iccv2017} we do not use more than 50 frames for the `Latest' setting. For each dataset and setting, we resample the input videos such that all sequences have a constant number of frames.  Due to unavailability of baseline results and following \cite{shi2018}, for UCF101-24 we report evaluate using 50\% of the frames from each video and for the TV-HI dataset, as in \cite{lan2014,gao2017red}, we consider only 1 seconds worth frames.

\subsubsection{Comparison to the state-of-the-art methods}

Evaluations for UCF101, UCF101-24, UTI and TV-HI datasets are presented in Tables \ref{tab:tab_1}, to \ref{tab:tab_4} respectively. Considering the results, the authors of Multi\_stage LSTM \cite{mohammad_iccv2017} and RED \cite{gao2017red} have introduced a new hand engineered loss that encourages the early prediction of the action class. The authors of RBF-RNN \cite{shi2018} use a GAN learning process where the loss function is also automatically learnt. Similar to the proposed architecture, the RBF-RNN \cite{shi2018} model also utilises the spatial representation of the scene through a Deep CNN model and tries to predict the future scene representations. However in contrast to the proposed architecture this method does not utilise temporal features, or joint learning. We learn a context descriptor which effectively combines both spatial and temporal representations which not only aids the action classification but also anticipates the future representations more accurately.  This led us to obtain superior results. In Tab. \ref{tab:tab_2}, the results for UCF101-24 shows that our model is able to outperform RBF-RNN \cite{shi2018} by 0.9\% while in Tab. \ref{tab:tab_3} we outperform \cite{shi2018} on the UTI dataset by 1.3\% at the earliest setting. %, where we demonstrate the importance of joint learning as well as utilising both visual and temporal features. 
%Please refer to Sec. \ref{sec:ablation} where we experimentally evaluate the contribution from joint learning of the model as well as individual features for the augmented performance of the proposed method.

When comparing the performance gap between the earliest and latest settings, our model has a smaller performance drop compared to the baseline models. The gap for UCF101 on our model is 1.4\% while the gap for the Multi\_stage LSTM model \cite{mohammad_iccv2017} is 2.9\%. $G^{V}$ and $G^{TP}$ synthesise the future representation of both visual and temporal streams while considering the current context. As such, the proposed model is able to better anticipate future actions, even with fewer frames. Our evaluations on multiple benchmarks further illustrate the generalisability of the proposed architecture, with varying video lengths and dataset sizes.    

\begin{table}[htbp]
\begin{center}
\resizebox{0.8\linewidth}{!}{
\begin{tabular}{c|c|c}
 \hline
 \hline
       \textbf{Method}  & \textbf{Earliest} & \textbf{Latest} \\	
 \hline
   	Context-aware $+$ loss in \cite{jain2016}& 30.6 & 71.1 \\	
	Context-aware $+$ loss in\cite{ma2016}& 22.6 & 73.1 \\
	Multi\_stage LSTM \cite{mohammad_iccv2017} & 80.5 & 83.4 \\
\hline
	\textbf{Proposed}& \textbf{84.2} & \textbf{85.6} \\ 
\hline
			
\end{tabular} }
\end{center}
\caption{Action anticipation results for UCF101 considering the `Earliest' 20\% of frames and `Latest' 50\% of frames.}\label{tab:tab_1}
\end{table}
%%%%%%%%%%%%%%%%%%%%%%%%

\begin{table}[htbp]
\begin{center}
\resizebox{0.6\linewidth}{!}{
\begin{tabular}{c|c}
 \hline
 \hline
       \textbf{Method}  & \textbf{Accuracy} \\	
 \hline
   	Temporal Fusion \cite{tempo_fusion}& 86.0 \\
	ROAD \cite{UCF101_24}& 92.0 \\
	ROAD + BroxFlow \cite{UCF101_24}& 90.0  \\
	RBF-RNN\cite{shi2018} & 98.0 \\
\hline
	\textbf{Proposed}& \textbf{98.9} \\
\hline
			
\end{tabular} }
\end{center}
\caption{Action anticipation results for UCF101-24 considering 50\% of frames from each video.}\label{tab:tab_2}
\end{table}
%%%%%%%%%%%%%%%%%%%%%%%%
\begin{table}[htbp]
\begin{center}
\resizebox{0.8\linewidth}{!}{
\begin{tabular}{c|c|c}
 \hline
 \hline
      \textbf{Method}  & \textbf{Earliest} & \textbf{Latest} \\	
 \hline
   	S-SVM \cite{soomro2018}& 11.0 & 13.4 \\
	DP-SVM \cite{soomro2018}& 13.0 & 14.6 \\
	CuboidBayes \cite{ryoo2011_anticipation1}& 25.0 & 71.7 \\
	CuboidSVM \cite{ryoo2010}& 31.7 & 85.0\\
	Context-aware$+$ loss in \cite{jain2016}& 45.0 & 65.0 \\
	Context-aware $+$ loss in\cite{ma2016}& 48.0 & 60.0 \\
	I-BoW \cite{ryoo2011_anticipation1}& 65.0 & 81.7 \\
	BP-SVM \cite{laviers2009}& 65.0 & 83.3 \\
	D-BoW \cite{ryoo2011_anticipation1}& 70.0 & 85.0\\
	multi-stageLSTM \cite{mohammad_iccv2017}& 84.0 & 90.0 \\
	Future-dynamic \cite{Rodriguez2018_eccvw}& 89.2 & 91.9 \\
	RBF-RNN \cite{shi2018}& 97.0 & NA\\
\hline
	\textbf{Proposed}& \textbf{98.3}  & \textbf{99.2} \\
\hline
			
\end{tabular} }
\end{center}
\caption{Action anticipation results for UTI `Earliest' 20\% of frames and `Latest' 50\% of frames. }\label{tab:tab_3}
\end{table}
%%%%%%%%%%%%%%%%%%%%%%%%

\begin{table}[htbp]
\begin{center}
\resizebox{0.6\linewidth}{!}{
\begin{tabular}{c|c}
 \hline
 \hline
       \textbf{Method}  & \textbf{Accuracy} \\	
 \hline
   	 Vondrick et. al \cite{vondrickCVPR16}& 43.6  \\
	RED \cite{gao2017red}& 50.2  \\	
\hline
	\textbf{Proposed}&  \textbf{55.7} \\
\hline
			
\end{tabular}}
\end{center}
\caption{Action anticipation results for TV Human Interaction dataset considering 1 second worth of frames from each video.}\label{tab:tab_4}
\end{table}

\subsection{Ablation Experiments}
\label{sec:ablation}
To further demonstrate the proposed AA-GAN method, we conducted an ablation study by strategically removing components of the proposed system. We evaluated seven non-GAN based model variants and ten GAN-based variants of the proposed AA-GAN model. Non-GAN based models are further broken into two categories: models with and  without future representation generators. Similarly, the GAN based models fall into two categories: those that do and do not learn tasks jointly Diagrams of these ablation models are available in the supplementary materials.

\textbf{Non-GAN based models:} These models do not utilise any future representation generators, and are only trained through classification loss. 

\begin{enumerate}[label=(\alph*)]
\item $\eta^{C, V}$: A model trained to classify using the context feature extracted only from the visual input stream (V).%a
\item $\eta^{C, TP}$: As per model (a), but using the temporal input stream (TP). %b
\item $\eta^{C, (V + TP)}$: As per (a), but using both data streams to create the context embedding. %c
\end{enumerate}

\textbf{Non-GAN based models with future representation generators:} Here, we add future embedding generators to the previous set of models. The generators are trained through mean squared error (i.e. no discriminator and no adversarial loss) while the classification is learnt through categorical cross entropy loss. The purpose of these models is to show how the joint learning can improve performance, and how a common embedding can serve both tasks.   

\begin{enumerate}[label=(\alph*)]
\setcounter{enumi}{3}   
\item $\eta^{C, V} + G^{V}$: Model with the future visual representation generator ($G^{V}$) and fed only with the visual input stream to train the classifier %d
\item $\eta^{C, TP} + G^{TP}$: As per (d), but receiving and predicting the temporal input stream. %e
\item $\eta^{C, (V + TP)} + G^{V} + G^{TP}$: The model is composed of both generators, $G^{V}$ and $G^{TP}$, and fed with both visual and temporal input streams.%f
\item $\eta^{C, (V + TP)} + G^{V} + G^{TP} +\mathrm{Att} $: As per (f) but with the use of attention to combine the streams. %g
\end{enumerate}

\textbf{GAN based models without joint training:} These methods are based on the GAN framework that generates future representations and a classifier that anticipates the action where these two tasks are learnt separately. We first train the GAN model using the adversarial loss and once this model is trained, using the generated future embeddings the classifier anticipates the action.  

\begin{enumerate}[label=(\alph*)]
\setcounter{enumi}{7}   
\item $\eta^{C, V} + GAN^{V}\backslash Joint $: Use the GAN learning framework with only the visual input stream and cosine distance based regularisation is used. %h
\item $\eta^{C, TP} + GAN^{TP}\backslash Joint$: As per (h), but with the temporal input stream %i
\item AA-GAN $\backslash Joint$ Use the GAN learning framework with both the visual and temporal input streams. %j
\end{enumerate}

\textbf{GAN based models with joint training:} These models train the deep future representation generators adversarially. The stated model variants are introduced by removing the different components from the proposed model.
  
\begin{enumerate}[label=(\alph*)]
\setcounter{enumi}{10}   
\item $\eta^{C, V} + GAN^{V}\backslash (\mathcal{L}^{R})$: The proposed approach with only the visual input stream and without cosine distance based regularisation. %k
\item $\eta^{C, TP} + GAN^{TP}\backslash (\mathcal{L}^{R})$: The proposed approach with only the temporal input stream and without cosine distance based regularisation. %l
\item $\eta^{C, V} + GAN^{V}$: The proposed approach with only the visual input stream. Cosine distance based regularisation is used.%m
\item $\eta^{C, TP} + GAN^{TP}$: The proposed approach with only the temporal input stream. Cosine distance based regularisation is used. %n
\item AA-GAN $\backslash (\mathcal{L}^{R})$ : Proposed model without cosine distance based regularisation. %o
\item AA-GAN $\backslash (DR)$ : Similar to the proposed model, however $G^{V}$ and $G^{TP}$ predict pixel values for future visual and temporal frames instead of representations extracted from the pre-trained feature extractor. %p
\end{enumerate}

\begin{table}[htbp]
\begin{center}
\resizebox{0.95\linewidth}{!}{
\begin{tabular}{c|c}
 \hline
 \hline
       \textbf{Method}  & \textbf{Accuracy} \\	
 \hline
   	 \rowcolor[HTML]{9ACBCC} 
	 (a) \hspace{1mm}  $\eta^{C, V}$ & 45.1 \\
	 \rowcolor[HTML]{9ACBCC} 
	(b) \hspace{1mm}  $\eta^{C, TP}$ & 39.8  \\
	\rowcolor[HTML]{9ACBCC} 
	(c) \hspace{1mm}  $\eta^{C, (V + TP)}$ & 52.0  \\
	 %\rule[-.002ex]{0pt}{0pt}    & \\	 	 	
%\hline 
 %         \rule[-.002ex]{0pt}{0pt}    & \\
        \rowcolor[HTML]{A7A8BA}
	(d) \hspace{1mm}  $\eta^{C, V} + G^{V}$ & 54.7 \\
        \rowcolor[HTML]{A7A8BA}
	(e) \hspace{1mm}  $\eta^{C, TP} + G^{TP}$ & 52.4  \\
	\rowcolor[HTML]{A7A8BA}
	(f)  \hspace{1mm} $\eta^{C, (V + TP)} + G^{V} + G^{TP}$ & 68.1\\
	\rowcolor[HTML]{A7A8BA}
	(g) \hspace{1mm}  $\eta^{C, (V + TP)} + G^{V} + G^{TP} +\mathrm{Att} $ & 68.8 \\
	% \rule[-.002ex]{0pt}{0pt}    & \\
%\hline
%	 \rule[-.002ex]{0pt}{0pt}    & \\
	 \rowcolor[HTML]{E1D4EE}
	(h) \hspace{1mm} $\eta^{C, V} + GAN^{V}\backslash Joint $ & 98.1 \\
	 \rowcolor[HTML]{E1D4EE}
	(i) \hspace{1mm} $\eta^{C, TP} + GAN^{TP}\backslash Joint$ & 97.9 \\
	\rowcolor[HTML]{E1D4EE}
	(j) \hspace{1mm} AA-GAN $\backslash Joint$ & 98.3 \\

%\hline
  	 %\rule[-.002ex]{0pt}{0pt}    & \\
	 \rowcolor[HTML]{F4E19E}
	(k)\hspace{1mm} $\eta^{C, V} + GAN^{V}\backslash (\mathcal{L}^{R})$ & 96.0 \\
	 \rowcolor[HTML]{F4E19E}
	 (l)\hspace{1mm} $\eta^{C, TP} + GAN^{TP}\backslash (\mathcal{L}^{R})$ & 95.4  \\
	\rowcolor[HTML]{F4E19E}
	(m) \hspace{1mm} $\eta^{C, V} + GAN^{V}$ & 98.4 \\
	\rowcolor[HTML]{F4E19E}
	(n) \hspace{1mm} $\eta^{C, TP} + GAN^{TP}$ & 98.1  \\
	%(p) \hspace{1mm}$\eta^{C, (V + TP)} + GAN^{V} + GAN^{TP}$ & 97.6 \\
	\rowcolor[HTML]{F4E19E}
	(o) \hspace{1mm} AA-GAN $\backslash (\mathcal{L}^{R})$ & 98.7 \\

 %\hline
%  	 \rule[-.002ex]{0pt}{0pt}    & \\
	 \rowcolor[HTML]{C9CAF4}
         (p) \hspace{1mm} AA-GAN $\backslash (DR)$  & 95.9\\
 %\hline
 %       \rule[-.002ex]{0pt}{0pt}    & \\
        \rowcolor[HTML]{F0BDBD}
	\textbf{AA-GAN (proposed)}&  98.9 \\
\hline
			
\end{tabular}}
\end{center}
\caption{Ablation results for UCF101-24 dataset for the `Latest' setting, which uses 50\% of the frames from each video.}\label{tab:tab_6}
\end{table}

\begin{figure}[h!]
  \centering
     \subfigure[AA-GAN]{\includegraphics[width=0.35\textwidth]{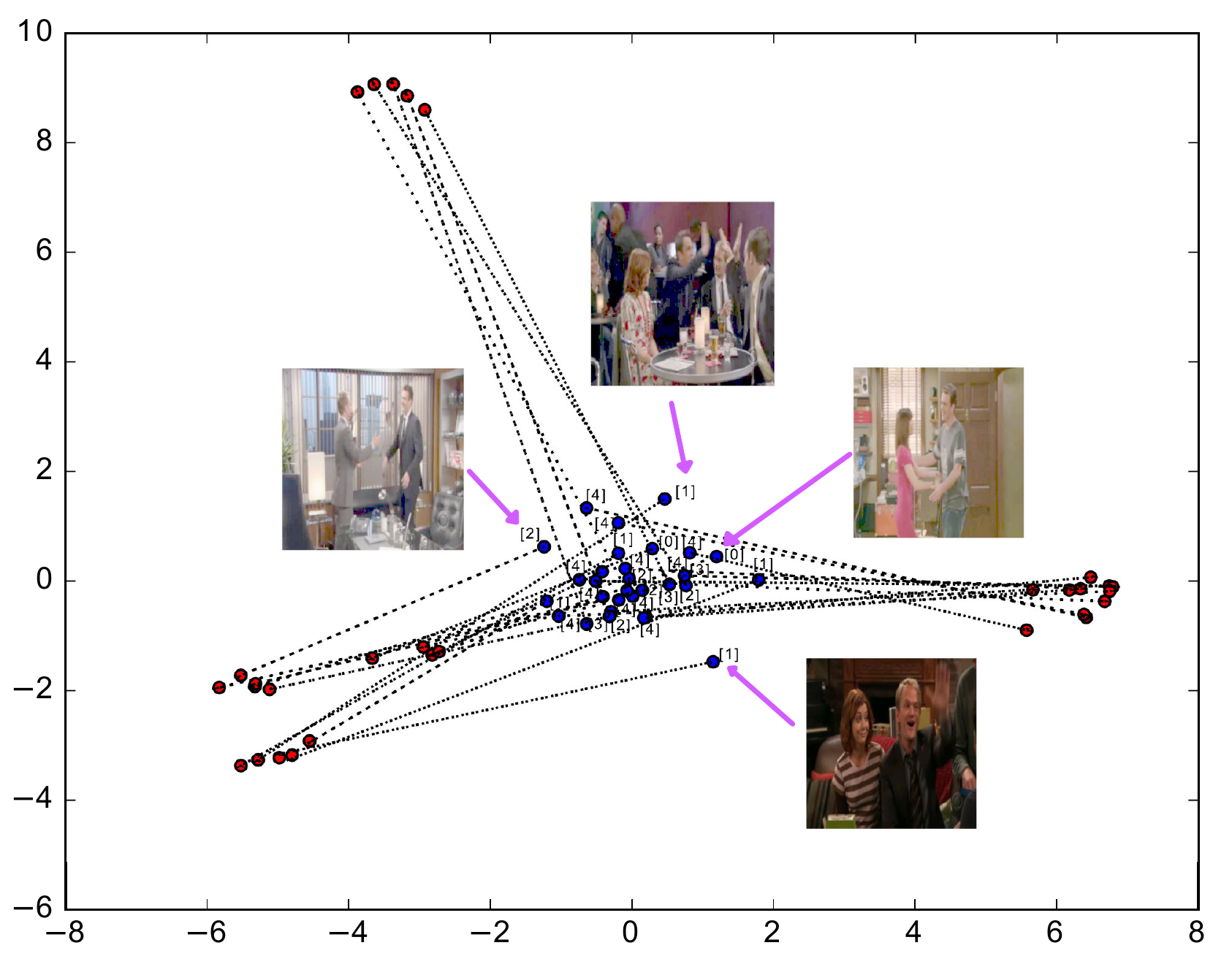}}
    \subfigure[Ablation model (g)(see Section \ref{sec:ablation})]{\includegraphics[width=0.35\textwidth]{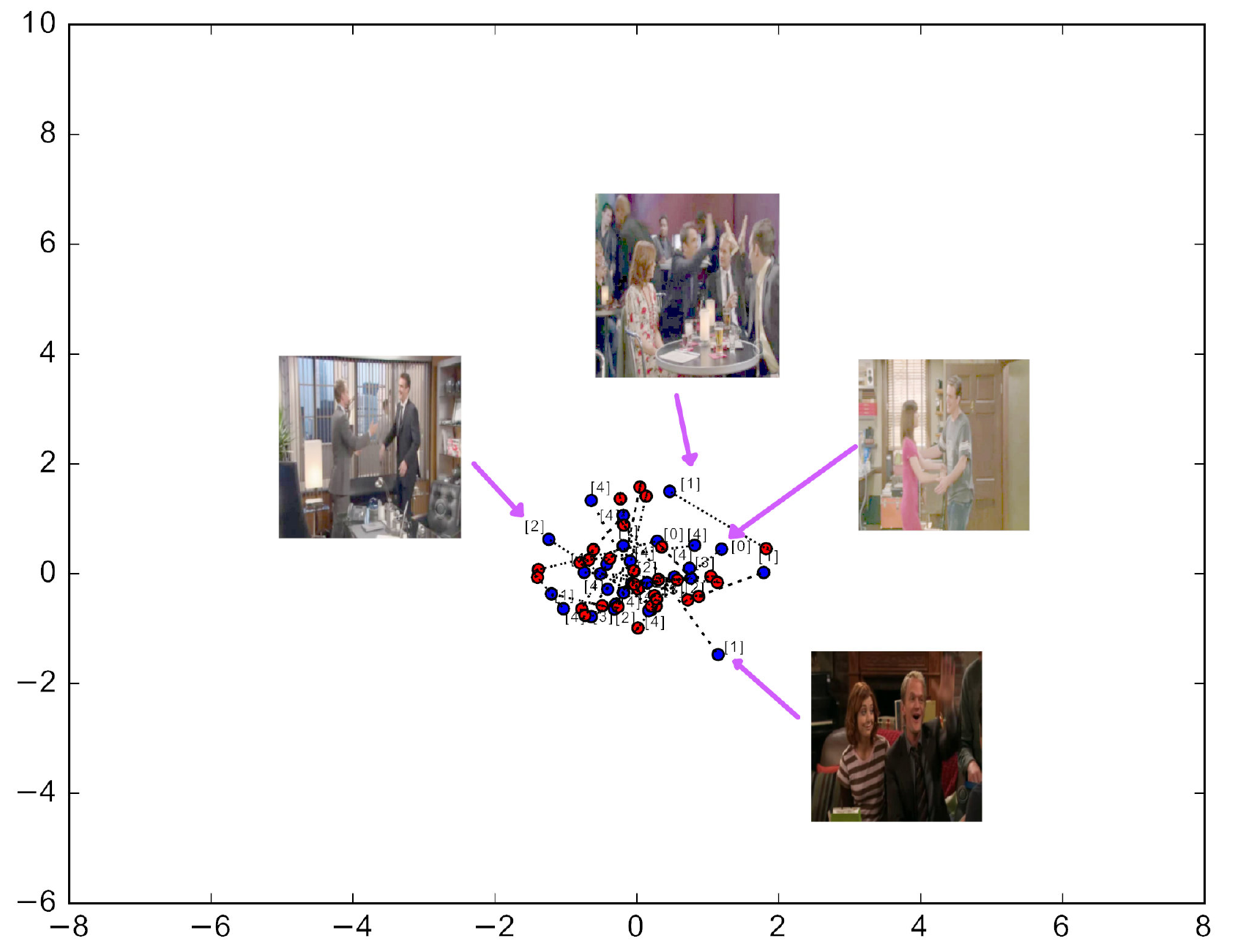}}
 \vspace{-1mm}
   \caption{ Projections of the discriminator hidden states for the for the AA-GAN (a) and ablation model (g) in (b) before (in blue) and after (in red) training. Ground truth action classes are in brackets. Insert indicates sample frames from the respective videos.}
  \label{fig:embedding_shift_AA_GAN}
\end{figure}

The evaluation results of the ablation models on the UCF101-24 test set are presented in Tab. \ref{tab:tab_6}. 

\textbf{Non-GAN based models (a to g):} Model performance clearly improves when using both data streams together over either one individually (see (c) vs (a) and (b); and (f) vs (d) and (e)). Hence, it is clear that both streams provide different information cues to facilitate the prediction. Comparing the results of models that do not utilise the future representation generators to (d), we see that overseeing future representation does improve the results.      

\textbf{GAN based models without joint training (h to j):} Comparing the non-GAN based methods with ablation model (h), we see that a major performance boost is achieved through the GAN learning process, denoting the importance of the automated loss function learning. Comparing the performance of visual and temporal streams, we observe that the visual stream is dominant, however combining both streams through the proposed attention mechanism captures complimentary information. 

\textbf{GAN based models with joint training (k to p):} Comparing models (h) and (i), which are single modal models that do not use joint training, with models (m) and (n) which do, we can see the clear benefit offered by learning the two complementary tasks together. This contradicts the observation reported in \cite{shi2018}, who use a classifier which was connected to the predicted future embeddings. We speculate that by learning a compressed context representation for both tasks we effectively propagate the effect of the action anticipation error through the encoding mechanisms, allowing this representation to be informative for both tasks.
Finally, by coupling the GAN loss together with $\mathcal{L}^{R}$, where the cosine distance based regularisation is combined with the exponential loss to encourage accurate long-term predictions, we achieve state-of-the-art results. Furthermore we compare the proposed AA-GAN model, where $G^V$ and $G^{TP}$ synthesise future visual and temporal representations, against ablation model (p) where $G^V$ and $G^{TP}$ synthesise pixel values for future frames. It is evident that the latter model fails to capture the semantic relationships between the low-level pixel features and the action class, leading to the derived context descriptor being less informative for action classification, reducing performance. 

%%%%%%%%%%

To demonstrate the discriminative nature of the learnt context embeddings, Fig. \ref{fig:embedding_shift_AA_GAN} (a) visualises the embedding space before (in blue) and after (in red) training of the proposed context descriptor for 30 randomly selected examples of the TV-HI test set. We extracted the learned context descriptor, $C_{t}$, and applied PCA \cite{wold1987principal} to generate 2D vectors. Ground truth action classes are indicated in brackets. 

This clearly shows that the proposed context descriptor learns embeddings which are informative for both future representation generation and the segregation of action classes. From the inserts which show sample frames from the videos, visual similarities exist between the classes, hence the overlap in the embedding space before training. However after learning, the context descriptor has been able to maximise the interclass distance while minimising the distance within the class. Fig. \ref{fig:embedding_shift_AA_GAN} (b) shows an equivalent plot for the ablation model (g). Given the cluttered nature of the embeddings before and after learning, it is clear that the proposed GAN learning process makes a significant contribution to learning discriminative embeddings~\footnote{Additional qualitative evaluations showing generated future visual and temporal representations are in the supplementary material.}

\subsection{Time Complexity}
We evaluate the computational demands of the proposed AA-GAN model for the UTI dataset's `Earliest' setting. The model contains 43M trainable parameters, and generates 500 predictions (including future visual and temporal predictions and the action anticipation) in 1.64 seconds using a single core of an Intel E5-2680 2.50 GHz CPU. 

\section{Conclusion}
In this paper we propose a framework which jointly learns to anticipate an action while also synthesising future scene embeddings. We learn a context descriptor which facilitates both of these tasks by systematically attending to individual input streams and effectively extracts the salient features. This method exhibits traits analogous to human neurological behaviour in synthesising the future, and renders an end to end learning platform. Additionally, we introduced a cosine distance based regularisation method to guide the generators in the synthesis task. Our evaluations demonstrate the superior performance of the proposed method on multiple public benchmarks.

{\small
\bibliographystyle{ieee_fullname}
\bibliography{paper6}
}

\end{document}